\documentclass[runningheads]{llncs}

\usepackage{eccv}
\usepackage{eccvabbrv}
\usepackage{algorithm}
\usepackage{algpseudocode}
\usepackage{subcaption}
\usepackage{graphicx}
\usepackage{booktabs}
\usepackage{pifont}
\usepackage[accsupp]{axessibility}
\usepackage{hyperref}

\newcommand{\cmark}{\ding{51}}%
\newcommand{\xmark}{\ding{55}}%

\begin{document}

\title{HSFM: Hard-Set-Guided Feature-Space Meta-Learning for Robust Classification under Spurious Correlations}

\titlerunning{HSFM: Feature-Space Meta-Learning for Spurious Correlations}

\author{
Aryan Yazdan Parast \and
Khawar Islam \and
Soyoun Won \and
Basim Azam \and
Naveed Akhtar
}

\institute{
The University of Melbourne, Melbourne, Australia
}
\authorrunning{A. Yazdan Parast et al.}

\maketitle
\begin{abstract}
Deep neural networks often rely on spurious features to make predictions, which makes them brittle under distribution shift and on samples where the spurious correlation does not hold (e.g., minority-group examples). Recent studies have shown that, even in such settings, the feature extractor of an Empirical Risk Minimization (ERM)-trained model can learn rich and informative representations, and that much of the failure may be attributed to the classifier head. In particular, retraining a lightweight head while keeping the backbone frozen can substantially improve performance on shifted distributions and minority groups. Motivated by this observation, we propose a bilevel meta-learning method that performs augmentation directly in feature space to improve spurious correlation handling in the classifier head. Our method learns support-side feature edits such that, after a small number of inner-loop updates on the edited features, the classifier achieves lower loss on hard examples and improved worst-group performance. By operating at the backbone output rather than in pixel space or through end-to-end optimization, the method is highly efficient and stable, requiring only a few minutes of training on a single GPU. We further validate our method with CLIP-based visualizations, showing that the learned feature-space updates induce semantically meaningful shifts aligned with spurious attributes.
\keywords{spurious correlations \and robust classification \and meta-learning \and feature-space augmentation}
\end{abstract}


\section{Introduction}
\label{sec:intro}

Training data in real-world datasets often contain spurious correlations between input features and the target label. Some features may appear predictive of the label in the training distribution, even though they are not causally related to it and may not remain predictive under distribution shift \cite{SalientImagenet, Fishermannet}. For example, in ImageNet, many images labeled as butterfly also contain flowers. Although flowers are not the cause of the butterfly label, their frequent co-occurrence can make them appear predictive during training \cite{SalientImagenet}. Prior work has shown that deep neural networks trained with empirical risk minimization (ERM) are vulnerable to shortcut learning, often relying on such spurious features to make predictions. As a result, their performance can degrade substantially on samples where these correlations do not hold \cite{IRM, recognita, NLPtendency, cnntendency}. This issue is especially critical in high-stakes domains such as medical diagnosis, where similar spurious patterns may also arise \cite{xray}.

A common approach to address this problem is to model the data distribution as a set of groups, where majority-group samples tend to exhibit the spurious correlation while minority-group samples do not. In this setting, robustness is typically evaluated by worst-group accuracy, and the objective is to improve performance on the worst-performing group \cite{jtt, GDRO, DFR}. It is now known that, although ERM classifiers often rely on spurious features when making predictions, their pretrained backbones can still encode both core and spurious cues, yielding informative feature representations \cite{DFR}. This suggests that the main source of failure may lie in the linear classifier built on top of these features rather than in the representation itself. Based on this insight, we propose HSFM - Hard-Set-Guided Feature-Space Meta-Learning for Robust Classification. HSFM optimizes support embeddings directly in feature space so that adapting the linear head on these learned samples reduces loss on hard validation examples, leading to improved robustness on minority-group samples.

By leveraging a bilevel optimization framework \cite{meta-learning}, our method extracts supervision from the failure modes of the linear head and uses it to optimize training representations directly in feature space. In contrast to several recent state-of-the-art approaches, HSFM does not rely on task-specific assumptions \cite{ddb, Masktune, dac}, explicit group annotations \cite{LISA, GDRO}, or auxiliary supervision. Compared with DFR \cite{DFR}, the key difference lies in how supervision is used: DFR retrains the linear head on a balanced validation subset, whereas HSFM explicitly identifies hard examples and uses their loss as a meta-optimization signal to update support embeddings in feature space. This allows HSFM to target the classifier’s evolving failure cases more directly.

Importantly, HSFM retains the same practical advantage that makes DFR appealing: it can be applied in only a few minutes on a single GPU. At the same time, we show that HSFM is not limited to spurious-correlation benchmarks and also improves performance on fine-grained classification tasks. In this setting, using only a pretrained ResNet backbone and a short optimization stage, HSFM can match or even outperform the ERM model, while DFR fails to achieve the same level of improvement. We further show that HSFM generalizes to alternative feature backbones, such as CLIP \cite{clip}. Finally, we demonstrate that the optimized embeddings can be used to guide a diffusion-based unCLIP generation process \cite{latent_diff}, providing a simple and interpretable way to expose and visualize model weaknesses through feature-space edits.

Together, these results show that HSFM provides a simple, efficient, and broadly applicable framework for improving robustness and analyzing model failure modes through feature-space optimization.

\section{Related Work}
Our work is most closely related to research on spurious correlation suppression, group-robust learning, and bilevel optimization. We briefly situate HSFM in these directions below.

\noindent \textbf{Group Robustness Under Spurious Correlation}. 
A broad line of work improves worst-group accuracy under spurious correlations and group imbalance. GroupDRO \cite{GDRO} directly minimizes the worst-group loss during training, but requires group labels. Other approaches instead reweight training samples \cite{jtt, DFR, afr, lff}. JTT \cite{jtt} retrains an ERM model with emphasis on samples misclassified by an initial ERM model. AFR \cite{afr} assigns weights using ERM-based incorrectness signals, while LfF \cite{lff} leverages an auxiliary biased model to upweight samples that are harder to learn. Similar to these methods, we use the model’s own error signal to identify hard examples. Unlike them, however, we do not only reweight existing samples; instead, we optimize representations to directly reduce loss on hard examples, enabling the method to move beyond the support of the original training set.

Another line of work mitigates spurious correlations through data augmentation followed by retraining \cite{Masktune, DISC, dac, ddb, ffr}. MaskTune \cite{Masktune} masks regions assumed to contain spurious cues and retrains on the masked images. DISC \cite{DISC} uses an auxiliary concept bank to identify spurious concepts and intervene accordingly. DaC \cite{dac} combines attribution-guided masking with image composition. DDB \cite{ddb} and FFR \cite{ffr} generate synthetic samples to improve coverage of underrepresented groups. In contrast, our method is driven directly by the optimization signal from hard examples, rather than relying on hand-crafted augmentation rules or external generative priors. Generator-based methods can incur heavy computational overhead and depend on the quality and controllability of the generated samples. Our method does not have such limitations. 

\noindent\textbf{Meta-learning and Bilevel Optimization}. 
Meta-learning, or `learning to learn', originates from early work on adaptive learning systems that sought to improve the learning process itself rather than only fit a fixed predictor \cite{EvolutionaryPI,Bengio1991LearningAS,metasurvey}. More recent methods formulate meta-learning as a bilevel optimization problem, where an inner adaptation step is optimized through an outer objective \cite{maml, metaSGD, imaml, metasurvey}. A representative example is Model-Agnostic Meta-Learning (MAML), which learns an initialization that can adapt to a new task in a few gradient steps \cite{maml}. Our method follows the same high-level structure: a linear classifier is adapted on a support set in the inner loop, while the support representations are optimized in the outer loop to reduce loss on a hard query set. However, unlike conventional meta-learning, our objective is not rapid adaptation across tasks, but robustness on the model’s current failure cases.
 
Our method is also connected to data condensation and dataset distillation through a shared bilevel optimization framework. These methods learn a compact synthetic set such that training on it approximates training on the original dataset \cite{dd,ddsurvey,dd2,dd3}. Among them, the closest line of work is robust dataset distillation, which also uses a bilevel-style procedure to learn a compact synthetic support set for subgroup robustness \cite{dddro}. HSFM differs in both objective and representation space. It does not aim to summarize the full training distribution or replace the dataset with a distilled substitute. Instead, it optimizes support embeddings directly in a frozen feature space to improve hard-example performance during classifier adaptation, which is both more efficient and more closely aligned with our objective.

\section{Problem Setup}
\label{sec:setup}

We consider a classification task under spurious correlations, where each example is a triplet
$(x_i, y_i, g_i)$ consisting of an input image $x_i$, a class label $y_i \in \mathcal{Y}$, and a group index $g_i \in \{0,\dots,G-1\}$. Each group corresponds to a specific combination of the target label and an environment attribute.

As a motivating example, consider the Waterbirds dataset \cite{GDRO}, where the task is to classify images as waterbird or landbird. In this dataset, the environment attribute is the image background, which can be either land (e.g., forest) or water (e.g., lake or sea). This yields four groups corresponding to all combinations of bird type and background:
landbirds on land, landbirds on water, waterbirds on land, and waterbirds on water.
The training set is strongly imbalanced across these groups: most landbirds appear on land backgrounds, and most waterbirds appear on water backgrounds. Consequently, landbird-on-water and waterbird-on-land samples form minority groups, while the other two groups are majority groups. This imbalance induces a spurious correlation between the class label and the background.

A classifier trained on Waterbirds with the standard ERM tends to rely on the background (a spurious feature) rather than the bird itself (the core feature), leading to poor performance on minority-group samples. To evaluate robustness, we use worst-group accuracy (WGA), defined for a classifier $f$ as
\begin{equation}
\mathrm{WGA}(f) = \min_{g \in \{0,\dots,G-1\}} \mathrm{Acc}_g(f),
\end{equation}
where $\mathrm{Acc}_g(f)$ denotes the accuracy of $f$ on group $g$. Addressing the spurious correlation problem, our objective is to improve robustness of the model by maximizing the worst-group accuracy.

\paragraph{ERM model decomposition.}
Following prior observations \cite{DFR}  that pretrained representations can remain informative even when ERM performs poorly on minority groups, we decompose the classification model into a feature extractor and a lightweight prediction head. Let
\begin{equation}
h_i = \phi(x_i) \in \mathbb{R}^d,
\end{equation}
where $\phi(\cdot)$ denotes a pretrained visual backbone (e.g., ResNet) and $h_i$ is the corresponding feature representation. The classifier $f$ is then written as
\begin{equation}
f(x_i) = \mathrm{softmax}(W h_i + b),
\end{equation}
where $(W,b)$ are the parameters of a linear classification head.

In our setting, we keep the backbone $\phi$ frozen and focus on improving robustness through optimization in the  feature space and adaptation of the prediction head. This design is motivated by the idea that, under spurious correlations, failure often arises from the decision rule learned on top of the features rather than a complete lack of useful information in the features themselves.

\section{Methodology}
\label{sec:method}

\subsection{Objective}
\label{sec:z}

\begin{figure}[t]
    \centering
    
    \includegraphics[width=\linewidth, ]{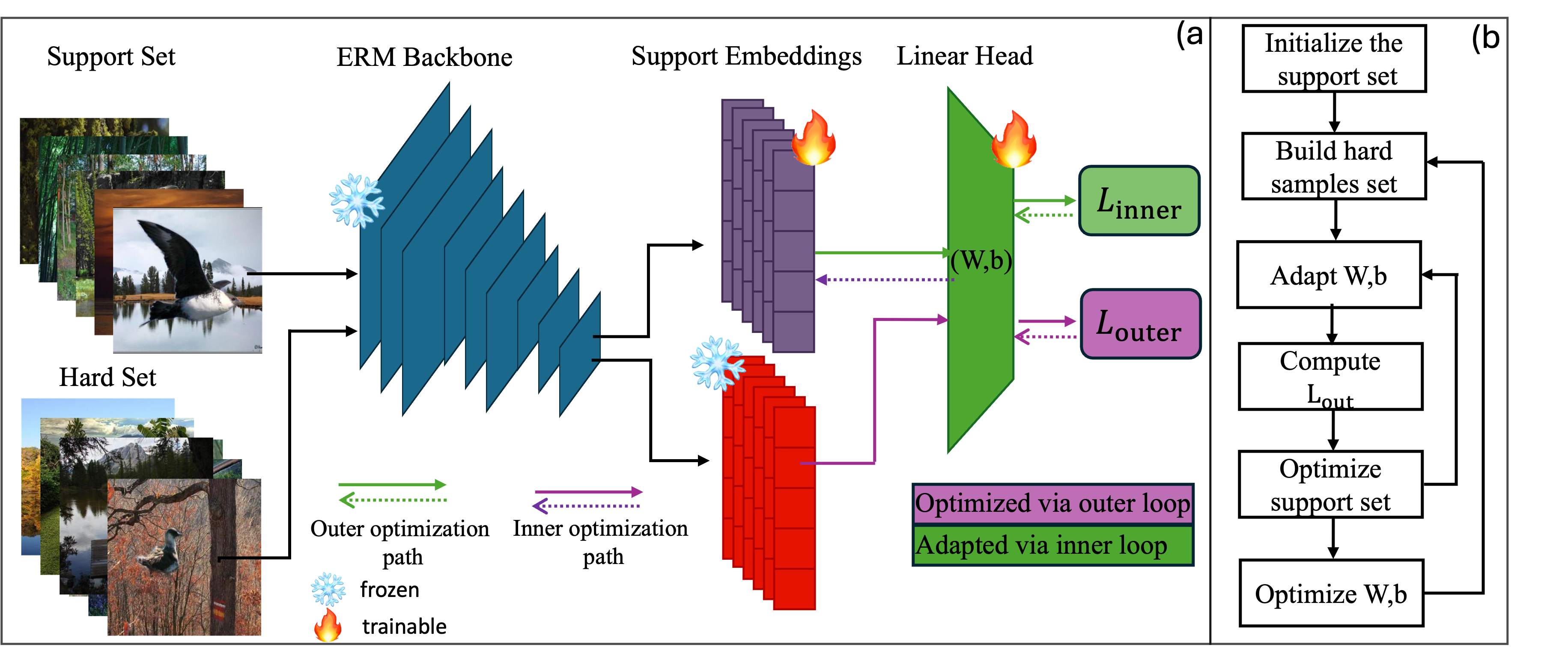}
    \caption{\textbf{Overview of HSFM.} \textbf{(a)} Initial training samples are passed through a frozen ERM feature extractor to obtain the support embeddings. The linear head is first adapted on the support set through the inner loss, and the outer loss is then computed on the hard set to improve performance on difficult examples. \textbf{(b)} Flowchart of the training procedure }
    \label{fig:main}
\end{figure}

ERM models trained on imbalanced datasets typically underperform on minority-group samples. As a consequence, these samples often incur higher classification loss than majority-group samples. In this work, without making additional assumptions about the dataset or the type of spurious correlation, we directly target this failure mode by focusing on reducing the loss on  high-loss examples.
For each validation sample $(x_i, y_i) \in \mathcal{D}_{\mathrm{val}}$, we define
\begin{equation}
\ell_i = \ell\big(f(x_i), y_i\big),
\end{equation}
where $\ell(\cdot,\cdot)$ is the cross-entropy loss. For each class $c \in \mathcal{Y}$, let

\begin{equation}
\label{topk}
\mathcal{Q}_c
=
\operatorname{TopK}\Big(\{(x_i,y_i)\in\mathcal{D}_{\mathrm{val}} \mid y_i=c\},\ K_{hard};\ \ell_i\Big),
\end{equation}
where $\operatorname{TopK}(\mathcal{A},K_{hard};\ell_i)$ returns the $K_{hard}$ elements of $\mathcal{A}$ with the largest loss values $\ell_i$.
 We then define the hard set as
\begin{equation}
\label{query}
\mathcal{Q} = \bigcup_{c \in \mathcal{Y}} \mathcal{Q}_c.
\end{equation}

Given the hard set $\mathcal{Q}$ defined above, let $\mathcal{S}$ denote a support set. Our goal is to optimize the support-side training variables such that the classifier obtained after training on $\mathcal{S}$ achieves low loss on the hard set $\mathcal{Q}$. Let $f'$ denote the adapted classifier obtained from $f$ after $T$ gradient-based update steps on $\mathcal{S}$:
\begin{equation}
f' = \mathrm{Adapt}(f,\mathcal{S};T).
\end{equation}
Equivalently, if $\theta$ denotes the parameters of the linear classifier, one adaptation step can be written as
\begin{equation}
\theta' = \theta - \alpha \nabla_{\theta}\mathcal{L}_{\mathrm{in}}(f_\theta),
\qquad f' = f_{\theta'},
\end{equation}
and similarly for multiple steps. The resulting adapted classifier is evaluated on the hard set using
\begin{equation}
\mathcal{L}_{\mathrm{hard}}(f';\mathcal{Q})
=
\frac{1}{|\mathcal{Q}|}\sum_{(x,y)\in\mathcal{Q}} \ell\!\left(f'(x),\, y\right),
\end{equation}
and we optimize the support-side variables so as to reduce this hard-set loss.

\subsection{Feature-Space Support Optimization}
\label{sec:feature_opt}

The support-side optimization in our method is performed in the representation space of a frozen ERM model. This choice is motivated by 
the idea that poor worst-group performance under ERM may not necessarily imply poor learned representations. ERM-trained backbones  encode multiple predictive cues \cite{DFR}, including both core object features and spurious background features, while the robustness failure may  stem from the linear decision rule learned on top of these representations. This makes representation-space intervention a promising choice in our setting. In addition, operating at the backbone output avoids repeated end-to-end optimization through a high-capacity network, making training substantially more stable and computationally efficient.

Let $\mathcal{S} \subset \mathcal{D}_{\mathrm{tr}}$ denote a support set used for adaptation. At the start of the training, we first sample a class-balanced subset of training examples and extract their frozen feature representations using the pretrained backbone:
\begin{equation}
h_i^{(0)} = \phi(x_i) \in \mathbb{R}^d, \qquad (x_i,y_i)\in\mathcal{S}.
\end{equation}
We then initialize a learnable support embedding $h_i \in \mathbb{R}^d$ with $h_i \leftarrow h_i^{(0)}$. 
Let
\begin{equation}
H = \{h_i \mid (x_i,y_i)\in\mathcal{S}\}
\end{equation}
denote the collection of optimized support embeddings. The adapted classifier $f'$ is obtained by training on the learned support set $\{(h_i, y_i)\}_{(x_i,y_i)\in\mathcal{S}}$, while $H$ is optimized so as to reduce the hard-set loss on $\mathcal{Q}$.

\subsection{Inner-Loop Adaptation}
\label{sec:inner}

Using the decomposition in Sec.~\ref{sec:setup}, we adapt only the linear head $(W,b)$ in the inner loop while keeping the backbone $\phi$ frozen. The inner-loop loss on the learned support embeddings is
\begin{equation}
\mathcal{L}_{\mathrm{in}}(W,b;H,\mathcal{S})
=
\frac{1}{|\mathcal{S}|}\sum_{(h_i,y_i)\in\mathcal{S}}
\ell\!\left(W h_i+b,\, y_i\right).
\end{equation}

Starting from initial head parameters $(W^{(0)}, b^{(0)})$, we perform $T$ gradient-based update steps:
\begin{equation}
(W^{(t+1)}, b^{(t+1)})
=
(W^{(t)}, b^{(t)})
-\alpha \nabla_{(W^{(t)}, b^{(t)})}
\mathcal{L}_{\mathrm{in}}(W^{(t)}, b^{(t)};H,\mathcal{S}),
\end{equation}
for $t=0,\dots,T-1$, where $\alpha$ is the inner-loop learning rate. We denote the adapted head by
\begin{equation}
(W', b') = (W^{(T)}, b^{(T)}).
\end{equation}

\subsection{Outer Objective on the Hard Set}
\label{sec:outer}

The adapted head is evaluated on the hard set $\mathcal{Q}$ defined in Sec.~\ref{sec:z}. For each $(x_j,y_j)\in\mathcal{Q}$, we compute the frozen hard feature
\begin{equation}
q_j = \phi(x_j),
\end{equation}
and define the outer hard-set loss as
\begin{equation}
\mathcal{L}_{\mathrm{out}}(H;\mathcal{Q})
=
\frac{1}{|\mathcal{Q}|}\sum_{(x_j,y_j)\in\mathcal{Q}}
\ell\!\left(W'(H)q_j+b'(H),\, y_j\right).
\end{equation}

The outer objective directly encourages the learned support embeddings $H$ to produce a classifier that performs better on the current set of high-loss validation examples.

\begin{algorithm}[t]
\caption{Hard-Set-Guided Feature-Space Meta-Learning}
\label{alg:feature_meta}
\begin{algorithmic}[1]
\State \textbf{Input:} $\mathcal{D}_{\mathrm{tr}}, \mathcal{D}_{\mathrm{val}}$, frozen backbone $\phi$, classifier head $(W,b)$
\State \textbf{Hyperparameters:} $K_{\mathrm{hard}}, K_H,  T$, epochs

    \State Sample a class-balanced support set $\mathcal{S}\subset\mathcal{D}_{\mathrm{tr}}$
    
    \State Initialize support embeddings $h_i \gets \phi(x_i)$ for $(x_i,y_i)\in\mathcal{S}$ and set $H=\{h_i\}$

    \For{iteration $=1$ to epochs}
        \State Build/refresh hard set $\mathcal{Q}$ from $\mathcal{D}_{\mathrm{val}}$ using top-$K_{\mathrm{hard}}$ loss samples per class under current classifier
        \For{$k=1$ to $K_H$}
            \State Compute $\mathcal{L}_{\mathrm{in}}(W,b;H,\mathcal{S})$ and perform $T$ inner adaptation steps to obtain $(W',b')$
            \State Compute $\mathcal{L}_{\mathrm{out}}(H;\mathcal{Q})$
            \State Update $H$ using meta-gradients $\nabla_H \mathcal{L}_{\mathrm{out}}(H;\mathcal{Q})$ 
        \EndFor
        \For {i $=1$ to $T$}
        \State Update the linear head $(W,b)$ on  $\{(h_i,y_i)\}_{(x_i,y_i)\in\mathcal{S}}$ 
        \EndFor
    \EndFor

\end{algorithmic}
\end{algorithm}

\subsection{Meta-Optimization}
\label{sec:meta}

We optimize the support embeddings $H$ using explicit meta-gradients by backpropagating through the inner-loop updates of the linear head:
\begin{equation}
\label{featureopt}
H \leftarrow H - \eta \nabla_{H}\mathcal{L}_{\mathrm{out}}(H;\mathcal{Q}),
\end{equation}
where $\eta$ is the meta learning rate. This yields a bilevel optimization procedure:
\begin{itemize}
    \item \textbf{Inner loop:} adapts the linear head $(W,b)$ on the learned support embeddings $h_i$, and 
    \item \textbf{Outer loop:} updates $H$ such that the adapted head achieves lower loss on the hard set $\mathcal{Q}$.
\end{itemize}

\subsection{Training Procedure}
\label{sec:training_procedure}

At the start of training, we sample a class-balanced support set $\mathcal{S}\subset\mathcal{D}_{\mathrm{tr}}$, extract frozen backbone features, and initialize the learnable support embeddings $H=\{h_i\}_{(x_i,y_i)\in\mathcal{S}}$ with $h_i\leftarrow \phi(x_i)$. We then iterate for a fixed number of epochs and alternate between updating the support embeddings and updating the linear head. In each epoch, we first refresh the hard set $\mathcal{Q}$ by selecting the top-$K_{\mathrm{hard}}$ highest-loss validation examples per class under the current classifier. Next, for $K_H$ steps, we update $H$ using the meta-objective in Eq.~\ref{featureopt}. Finally, for $T$ steps, we update the classifier head $(W,b)$ on the current support embeddings. This alternating procedure jointly refines the support embeddings and the head, allowing the method to track and reduce the classifier's evolving failure cases. The full procedure is summarized in Algorithm~\ref{alg:feature_meta} and illustrated in Fig.~\ref{fig:main}.

\section{Experiments}
\label{sec:experiments}

We evaluate our method from four complementary perspectives. First, we assess its effectiveness on standard spurious correlation benchmarks, including \textbf{Waterbirds} \cite{Birds, GDRO}, \textbf{CelebA}\cite{celeba}, \textbf{Dominoes} \cite{dominoes}, and \textbf{MetaShift}\cite{metashift}, where the primary goal is to improve robustness under distribution shift and reduce performance degradation on minority groups. These experiments measure whether our feature-space meta-learning framework can consistently improve worst-group performance in settings where ERM models are known to rely on spurious cues.

Second, we study the broader applicability of our method on fine-grained classification tasks, using the \textbf{Stanford Cars} \cite{Cars},  Caltech-UCSD Birds-200-2011 \textbf{CUB}\cite{WahCUB_200_2011}, and \textbf{Oxford Flowers}\cite{Flowers} datasets. Unlike the spurious-correlation benchmarks, these datasets do not explicitly define majority and minority groups through known shortcut attributes. Instead, they present challenging visually similar categories, allowing us to examine whether our hard-set-guided optimization strategy can also improve generalization in fine-grained recognition settings by focusing training on difficult examples.

Third, we demonstrate that our method can be used as a model analysis tool to expose and visualize learned biases. In particular, we show how the optimized feature edits reveal which failure modes the classifier is most sensitive to, and how these directions can be paired with a diffusion-based unCLIP \cite{latent_diff, unclip} model to automatically generate visualizations that make the model's biases more interpretable. This provides a qualitative perspective on the behavior of the learned classifier beyond standard accuracy metrics.

Finally, we present an ablation study to analyze the contribution of the main components. We will include our hyperparameters and the efficiency discussion in the Appendix.

\subsection{Spurious Correlation}

\vspace{1mm}
\noindent\textbf{Datasets:}
We first briefly describe the datasets and the spurious correlations present in  them. More details on the datasets are available in the Appendix.

\vspace{1mm}
\noindent\textit{Waterbirds. \cite{GDRO}}
Waterbirds is a synthetic dataset constructed by placing bird images from CUB \cite{Birds} onto background scenes from Places \cite{places}, such as lakes, oceans, bamboo forests, and broadleaf forests. In this dataset, a spurious correlation exists between the background and the class label: most waterbirds appear on water backgrounds, while most landbirds appear on land backgrounds.

\vspace{1mm}
\noindent\textit{CelebA. \cite{celeba}}
CelebA contains celebrity face images, and the task is to classify hair color as blond or non-blond. In this setting, there is a spurious correlation between gender and hair color, as most blond celebrities are women. We follow the experimental setting of DaC \cite{dac} for this dataset.

\vspace{1mm}
\noindent\textit{Dominoes. \cite{dominoes}}
Dominoes is a synthetic dataset introduced in \cite{dominoes}. Each image is composed of two parts: a CIFAR-10 image (either truck or car), which determines the class label, and an MNIST digit (either 0 or 1), which serves as the spurious attribute. The MNIST digit is placed below the CIFAR image and is correlated with the target label.

\vspace{1mm}
\noindent\textit{MetaShift. \cite{metashift}}
MetaShift is a distribution-shift benchmark for classifying images of cats and dogs. Following \cite{DISC}, we use images of cats on beds and sofas, and dogs on benches and bikes, as the training set. We then evaluate the model on cat and dog images with shelf backgrounds in order to assess robustness under distribution shift. More details on the datasets are available in the Appendix.

\vspace{1mm}
\noindent\textbf{Models:}
For fair comparison, we use an ImageNet-pretrained ResNet-50 \cite{resnet50,imagenet} as the backbone, matching the baselines in Table~\ref{tab:method_comparison}. We first train it on each dataset with standard ERM, then use the resulting model to initialize HSFM. During HSFM training, we freeze the backbone and update only the linear head (Sec.~\ref{sec:method}). We repeat the same protocol with ViT-B/16 \cite{vitb} (Table~\ref{tab:vit-b}) and ConvNet \cite{liu2022convnet2020s} (Table ~\ref{tab:convex_backbone}). For CLIP ViT-H \cite{clip}, we train only a linear head for ERM to preserve the CLIP embedding space required for visualization (Sec.~\ref{sec:clip}).

\vspace{1mm}
\noindent\textbf{Results:}
We compare our method against several state-of-the-art approaches, including DFR \cite{DFR}, GroupDRO \cite{GDRO}, LISA \cite{LISA}, JTT \cite{jtt}, DaC \cite{dac}, and DDB \cite{ddb}. For DFR \cite{DFR}, GroupDRO\cite{GDRO}, LISA \cite{LISA}, JTT \cite{jtt}, and DaC \cite{dac}, we use the results reported by DaC \cite{dac}, as we follow the same experimental settings. For DDB \cite{ddb}, we report the results provided in the original paper.

We evaluate performance using both average accuracy and worst-group accuracy, with worst-group accuracy being the primary metric for spurious-correlation benchmarks and the standard measure of robustness in this setting. The results are summarized in Table~\ref{tab:method_comparison}.
Following \cite{DFR}, the Group Info column in Table~\ref{tab:method_comparison} indicates how each method uses group annotations. DFR \cite{DFR} uses validation group labels directly during training, which we denote by \cmark\cmark. In contrast, the other methods use validation group labels only for model selection.
\begin{table*}[ht!]
\caption{Comparison of worst-group accuracy and average accuracy between our method and the baselines. We report the mean and standard deviation over three runs with different random seeds. The best result overall is shown in bold, while the best result among methods that do not use group labels during training is underlined.}
\label{tab:method_comparison}
\centering
\resizebox{\textwidth}{!}{%
\small
\begin{tabular}{lccccccccc}
\toprule
& {Group~Info} & \multicolumn{2}{c}{Waterbirds} & \multicolumn{2}{c}{CelebA} & \multicolumn{2}{c}{MetaShift} & \multicolumn{2}{c}{Dominoes} \\ 
\cmidrule(lr){3-4} \cmidrule(lr){5-6} \cmidrule(lr){7-8} \cmidrule(lr){9-10}
Method & train/val & Worst & Average & Worst & Average & Worst & Average & Worst & Average \\ 
\midrule
DFR & \xmark/\cmark\cmark & $92.3_{\pm 0.2}$ & $93.3_{\pm 0.5}$ & $88.3_{\pm 1.1}$ & $91.3_{\pm 0.3}$ & $72.8_{\pm 0.6}$ & $77.5_{\pm 0.6}$ & $90.0_{\pm 0.4}$ & $92.3_{\pm 0.2}$ \\
Group DRO & \cmark/\cmark & $91.4_{\pm 1.1}$ & $93.5_{\pm 0.3}$ & $88.9_{\pm 2.3}$ & $92.9_{\pm 0.2}$ & $66.0_{\pm 3.8}$ & $73.6_{\pm 2.1}$ & - & - \\
LISA & \cmark/\cmark & $89.2_{\pm 0.6}$ & $91.8_{\pm 0.3}$ & $\boldsymbol{89.3_{\pm 1.1}}$ & $92.4_{\pm 0.4}$ & $59.8_{\pm 2.3}$ & $70.0_{\pm 0.7}$ & - & - \\
JTT & \xmark/\cmark & $86.7$ & $93.3$ & $81.1$ & $88.0$ & $64.6_{\pm 2.3}$ & $74.4_{\pm 0.6}$ & - & - \\
DaC & \xmark/\cmark & $92.3_{\pm 0.4}$ & $\underline{\boldsymbol{95.3_{\pm 0.4}}}$ & $81.9_{\pm 0.7}$ & $91.4_{\pm 1.1}$ & $78.3_{\pm 1.6}$ & $79.3_{\pm 0.1}$ & $89.2_{\pm 0.1}$ & $92.2_{\pm 0.3}$ \\
DDB & \xmark/\cmark & $93.0_{\pm 0.1}$ & $93.6_{\pm 0.1}$ & $85.8_{\pm 1.4}$ & $87.3_{\pm 0.7}$ & $\underline{\boldsymbol{81.2_{\pm 0.2}}}$ & $\underline{\boldsymbol{81.3_{\pm 0.2}}}$ & - & - \\
\midrule
Base (ERM) & \xmark/\xmark & $74.6$ & $90.21$ & $30.6$ & $\underline{\boldsymbol{95.83}}$ & $64.1$ & $75.65$ & $78.6$ & $88.83$ \\
HSFM (ours) & \xmark/\cmark & $\underline{\boldsymbol{93.1_{\pm 0.1}}}$ & $94.0_{\pm 0.5}$ & $\underline{89.2_{\pm 0.2}}$ & $90.6_{\pm 0.5}$ & $77.2_{\pm 0.1}$ & $77.4_{\pm 0.0}$ & $\underline{\boldsymbol{90.4_{\pm 0.3}}}$ & $\underline{\boldsymbol{92.9_{\pm 0.2}}}$ \\
\bottomrule
\end{tabular}
}
\end{table*}

\begin{table*}[t]
\caption{Comparison of worst-group accuracy and average between HSFM and ERM on four benchmarks with Vit-b-16 as backbone.}
\label{tab:vit-b}
\centering
\small
\resizebox{\textwidth}{!}{%
\begin{tabular}{lcccccccc}
\toprule
 & \multicolumn{2}{c}{Waterbirds} & \multicolumn{2}{c}{CelebA} & \multicolumn{2}{c}{MetaShift} & \multicolumn{2}{c}{Dominoes} \\
\cmidrule(lr){2-3}\cmidrule(lr){4-5}\cmidrule(lr){6-7}\cmidrule(lr){8-9}
Method & Worst & Average & Worst & Average & Worst & Average & Worst & Average \\
\midrule
Base (ERM)
& $53.9$ & $75.37$
& $53.3$ & $\mathbf{95.70}$
& $69.7$ & $69.78$
& $46.1$ & $70.20$ \\
HSFM (ours)
& $\mathbf{81.5}$ & $\mathbf{83.02}$
& $\mathbf{88.3}$ & $90.44$
& $\mathbf{69.9}$ & $\mathbf{70.22}$
& $\mathbf{73.3}$ & $\mathbf{75.46}$ \\
\bottomrule
\end{tabular}
}
\end{table*}

As reported in Table~\ref{tab:method_comparison}, HSFM achieves the best performance on Waterbirds and Dominoes. On CelebA, our method also outperforms all methods that do not use group labels during training. These results suggest that HSFM is broadly applicable across different spurious-correlation settings and does not suffer from the limitations of data-generation-based approaches such as DDB, which can struggle to produce suitable samples on synthetic or carefully constructed datasets such as Dominoes. More details are available in Appendix.

MetaShift is substantially smaller than the other benchmarks, with only 81 validation samples. As a result, DFR performs less reliably on this dataset, likely because it depends more directly on validation-set as their training set. In contrast, HSFM is able to outperform DFR by a clear margin, as it uses the validation set to guide feature-space augmentation through the hard-set objective. DDB and DaC achieve stronger results on MetaShift, which is expected, since they are data-augmentation-based methods and are less dependent on the small validation set in the same way.
Nevertheless, our result on Metashift shows that our method is effective under distribution shift and can improve robustness in that setting as well. Additional experiments in Section~\ref{fine-grained} further evaluate our method against DFR, which we treat as our primary baseline.

In Table~\ref{tab:vit-b}, we report results with a ViT-B backbone and in Table~\ref{tab:convex_backbone} with a ConvNet \cite{liu2022convnet2020s} backbone. HSFM consistently improves worst-group accuracy on Waterbirds, CelebA, and Dominoes, demonstrating robustness across backbones. On MetaShift, the gain is marginal, which we attribute to the limited validation set size.

\subsection{Fine-grained Classification}
\label{fine-grained}

We further evaluate HSFM on fine-grained benchmarks including Stanford-Cars \cite{Cars}, CUB-Birds \cite{WahCUB_200_2011}, and Oxford-Flowers102 \cite{Flowers} to assess whether our hard-set-guided feature space optimization generalizes beyond spurious correlation settings.
Unlike group-robustness benchmarks, these datasets involve highly similar categories, making them a natural testbed for evaluating whether targeting high-loss \textit{hard} examples can improve generalization.

\paragraph{Evaluation protocol.}
We consider two backbone initialization settings. \textbf{ERM} uses a ResNet-50 backbone that is first fine-tuned on the dataset training split with standard ERM, after which we freeze the backbone and apply HSFM to optimize support embeddings and adapt the linear head. \textbf{Pretrained} uses an ImageNet-pretrained ResNet-50 backbone without any dataset-specific fine-tuning; we directly freeze the backbone and apply HSFM on top of its features. This setting tests our hypothesis that a strong pretrained representation can already contain informative cues for fine-grained recognition, and that a large portion of the remaining error can be addressed by refining the linear decision rule via hard-set-guided support embedding optimization. In both settings, we report top-1 accuracy and compare against ERM and DFR baselines.

\paragraph{Results.}
Table~\ref{tab:fine_grained} reports top-1 accuracy. HSFM achieves consistent gains on Stanford Cars and Oxford-Flowers102. On Stanford Cars, HSFM-Pretrained attains the best accuracy (85.10\%), outperforming both ERM (83.98\%) and DFR baselines. On Oxford-Flowers102, HSFM-ERM achieves the best result (97.27\%), while HSFM-Pretrained reaches comparable performance (97.10\%), indicating that HSFM can be effective even without dataset-specific backbone fine-tuning. 
On CUB-Birds, ERM attains the best accuracy (75.01\%). However, HSFM-Pretrained remains competitive (72.94\%) despite using only an ImageNet-pretrained frozen backbone and a lightweight post hoc optimization stage, within a few minutes. We provide detailed runtime and efficiency comparisons in the Appendix. Importantly, this behavior is specific to HSFM: while DFR is similarly efficient, it degrades substantially in the Pretrained setting, suggesting that simply re-fitting a linear head is insufficient. In contrast, HSFM remains effective in these fine-grained regimes.
\begin{table}[t]
\setlength{\tabcolsep}{6pt}
\centering
\caption{Comparison of top-1 classification accuracy on fine-grained classification benchmarks. The best result in each column is shown in bold.}
\label{tab:fine_grained}
\begin{tabular}{lccc}
\toprule
Method & Stanford Cars & CUB-Birds  & Oxford Flowers \\ 
\midrule
ERM & 83.98 & \textbf{75.01}  & 91.07  \\
DFR-ERM &  82.74  &  72.90 & 95.98  \\
DFR-Pretrained & 32.75  & 48.33  & 87.71  \\
\midrule
HSFM-ERM & 83.44 & 72.83 & \textbf{97.27}  \\
HSFM-Pretrained & \textbf{85.10}  & 72.94  & 97.10  \\
\bottomrule
\end{tabular}
\end{table}

\subsection{unCLIP Visualization}
\label{sec:clip}
\begin{figure}[t]
    \centering
    \includegraphics[width=\linewidth]{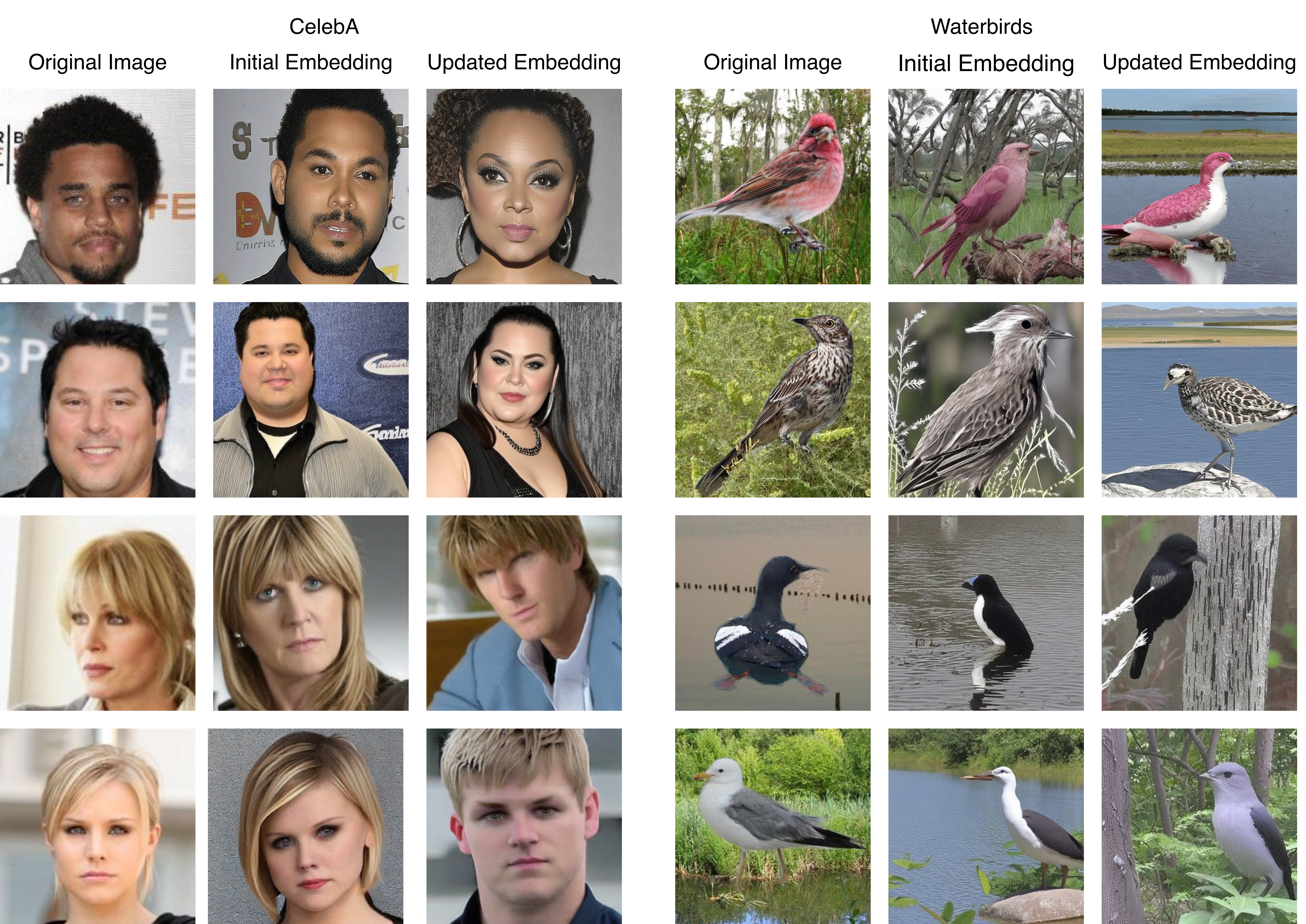}
    \caption{Visualization of three variants for each sample:  original image, the image generated by SD unCLIP from the initial embedding, and the image generated from the optimized embedding. In CelebA, the first two rows show non-blond males  shifted toward non-blond females. The last two rows show the shift from blond females to blond males. In Waterbirds, the first two rows show landbirds shifted from land backgrounds to water backgrounds, while the last two rows show waterbirds shifted from water backgrounds to land backgrounds.}
    \label{fig:unclip}
\end{figure}
\captionsetup[table]{skip=4pt}

\begin{table}[t]
\centering

\begin{subtable}[t]{0.49\linewidth}
\centering
\caption{Comparison of worst-group accuracy and average between HSFM and ERM with ConvexNet as backbone}
\label{tab:convex_backbone}
\begin{tabular}{lcccc}
\toprule
& \multicolumn{2}{c}{Waterbirds} & \multicolumn{2}{c}{CelebA} \\
\cmidrule(lr){2-3} \cmidrule(lr){4-5}
Method & Worst & Average & Worst & Average \\
\midrule
ERM  & $82.7$ & $90.99$ & $47.2$ & $\boldsymbol{96.04}$ \\
HSFM & $\boldsymbol{92.6}$ & $\boldsymbol{92.96}$ & $\boldsymbol{81.1}$ & $93.82$ \\
\bottomrule
\end{tabular}
\end{subtable}\hfill
\begin{subtable}[t]{0.49\linewidth}
\centering
\caption{Comparison of CLIP-based classifiers on the test set. We report average accuracy and worst-group accuracy for the ERM baseline and HSFM.}
\label{tab:clip_backbone}
\begin{tabular}{lcccc}
\toprule
& \multicolumn{2}{c}{Waterbirds} & \multicolumn{2}{c}{CelebA*} \\
\cmidrule(lr){2-3} \cmidrule(lr){4-5}
Method & Worst & Average & Worst & Average \\
\midrule
ERM  & $68.4$ & $89.39$ & $50.0$ & $74.50$ \\
HSFM  & $\boldsymbol{80.5}$ & $\boldsymbol{94.05}$ & $\boldsymbol{81.0}$ & $\boldsymbol{89.75}$ \\
\bottomrule
\end{tabular}
\end{subtable}

\end{table}
In this section, we demonstrate how our method can be used as a systematic tool for exposing model biases and failure modes. To this end, we train a linear classifier on top of the CLIP  vision encoder \cite{clip} for Waterbirds \cite{GDRO, WahCUB_200_2011} and CelebA* \cite{celeba}. For CelebA*, we use an adjusted version of the dataset with amplified spurious correlations in order to make the effect more visible; additional details are provided in the Appendix. Table~\ref{tab:clip_backbone} compares the baseline with our method, showing that HSFM generalizes to CLIP-based representations and continues to improve worst-group accuracy in this setting.

To visualize the optimized feature edits, we use Stable Diffusion (SD) unCLIP \cite{latent_diff, unclip}. SD unCLIP uses a CLIP image embedding to guide the diffusion process. Since our method updates image embeddings as in Eq.~\ref{featureopt}, we can visualize both the original embeddings and their optimized versions through image generation. Figure~\ref{fig:unclip} presents qualitative examples of these generated samples.

On Waterbirds, majority-group samples such as waterbirds on water backgrounds and landbirds on land backgrounds are transformed into waterbirds on land backgrounds and landbirds on water backgrounds, which correspond to minority-group configurations. We observe a similar trend on CelebA*, where majority-group samples are shifted toward minority-group configurations under our optimization. These results suggest that the learned feature updates are semantically aligned with the underlying dataset biases and can therefore be used to reveal the model's reliance on spurious attributes.

\subsection{Ablation Study}

\begin{figure}[t]
    \centering
    
    \begin{minipage}[t]{0.48\linewidth}
        \centering
        \includegraphics[width=\linewidth]{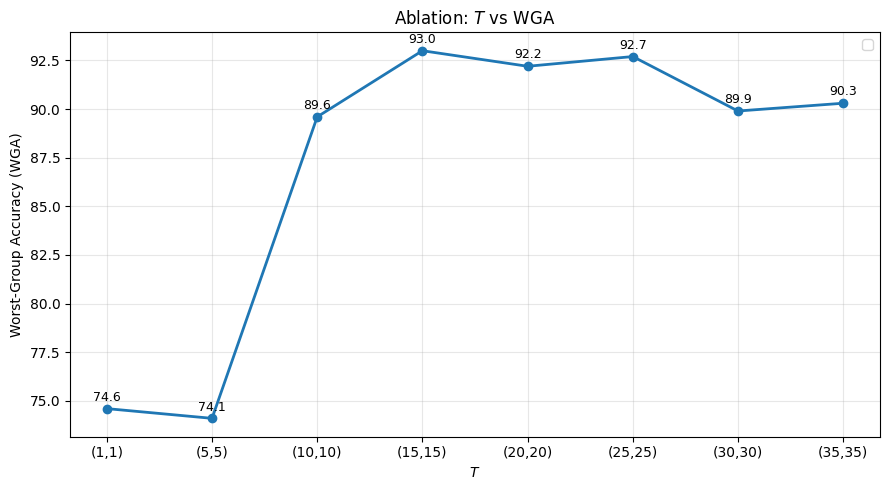}
        \captionof{figure}{{Worst-group accuracy (WGA) under different values of $T$ on the Waterbirds dataset.} }
        \label{fig:ablation_t_kf}
    \end{minipage}
    \hfill
    \begin{minipage}[t]{0.48\linewidth}
        \centering
        \includegraphics[width=\linewidth]{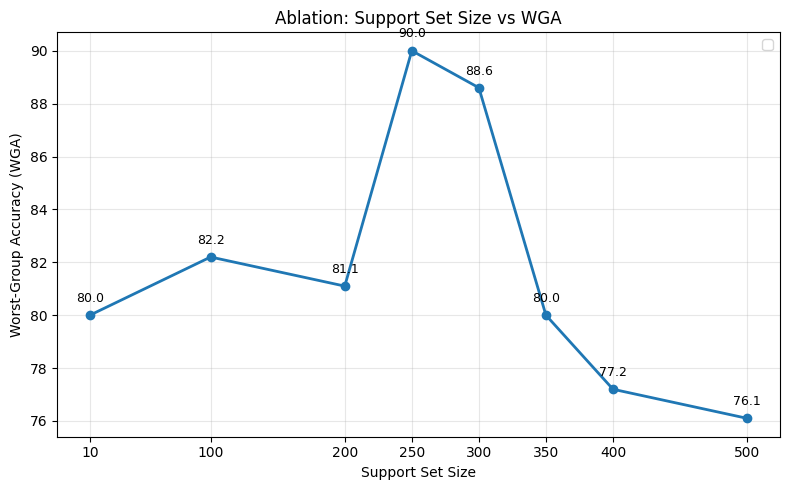}
        \captionof{figure}{{Worst-group accuracy (WGA) under different support set sizes on the CelebA dataset.} }
        \label{fig:ablation_support}
    \end{minipage}
\end{figure}

For the ablation studies, we report the worst-group accuracy (WGA) under different values of $T$ in Fig.~\ref{fig:ablation_t_kf}, and under different support set sizes in Fig.~\ref{fig:ablation_support}. In Fig.~\ref{fig:ablation_t_kf}, $T$ denotes the number of inner-loop updates and the number of ERM updates. Our method remains robust across a wide range of moderate and large values, achieving consistently high WGA for $T \geq 10$. However, when $T$ is very small (i.e., $1$ or $5$), performance degrades sharply, indicating that too few adaptation and ERM updates are insufficient to produce a reliable optimization signal.

In Fig.~\ref{fig:ablation_support}, the support set size controls the number of training samples used to initialize the learnable support embeddings. The best performance is obtained at an intermediate support set size, peaking at 250 samples. When the support set is too small, the learned embeddings may not capture sufficient diversity from the training distribution. In contrast, excessively large support sets can weaken the optimization focus by spreading the meta-objective over a broader and more heterogeneous set of samples. In particular, as the support set becomes larger, the optimization signal from the minority-group samples, can become diluted by the inclusion of easier samples. These results suggest that a moderate support set size provides the best trade-off between diversity and targeted feature-space optimization. Notably, even with very small support sets, HSFM still improves WGA, further demonstrating the robustness of the proposed method.

\section{Conclusion}
We introduced HSFM, a simple and effective method for improving classifier robustness by leveraging the model’s own signal on hard samples. Our results show that, through feature-space meta-learning, HSFM consistently improves worst-group accuracy on standard spurious-correlation benchmarks while also showing potential in fine-grained classification tasks. Beyond improving performance, we further show that the method can help visualize model weaknesses, offering additional insight into classifier failure modes.
\textbf{Limitations}. Our method relies on a frozen backbone producing sufficiently informative features. When this holds, optimizing in feature space can improve robustness by updating only the classifier. If not, gains may require fine-tuning the backbone, increasing computational cost and reducing the efficiency advantage of our approach. Notably though, the assumption is practical and holds in practice. 
\section{Datasets}
\subsection{Waterbirds}
Waterbirds is a synthetic dataset created by combining bird images from CUB \cite{Birds} with background scenes from Places \cite{places}. It introduces a spurious correlation between the bird class and the background: most waterbirds appear on water backgrounds, while most landbirds appear on land backgrounds. As shown in Table~\ref{tab:waterbirds_samples}, the dataset is also strongly imbalanced across groups, with far fewer minority-group samples than majority-group samples in the training set.

\begin{table}[ht!]
\caption{The number of samples in the training, validation, and test sets for the Waterbirds dataset.}
\label{tab:waterbirds_samples}
\centering
\small
\begin{tabular}{c|ccc}
\toprule
Split & Train & Validation & Test \\
\midrule
(landbird, land background) & 3,498 & 467 & 2,255 \\
\midrule
(landbird, water background) & 184 & 466 & 2,255 \\ 
\midrule
(waterbird, land background) & 56 & 133 & 642 \\
\midrule
(waterbird, water background) & 1,018 & 133 & 642 \\
\bottomrule
\end{tabular}
\end{table}
\subsection{CelebA}
CelebA is a face dataset of celebrity images, where the task is to classify hair color as \emph{blond} or \emph{non-blond}. In this setting, there is a spurious correlation between gender and hair color, since most blond celebrities are female. Following DaC \cite{dac}, we use the balanced class split between blond and non-blond shown in Table~\ref{tab:celebA_samples}. Despite this class balancing, the dataset remains highly imbalanced across groups, with \emph{(blond, male)} forming a clear minority group. This makes CelebA a standard benchmark for evaluating robustness under spurious correlations.

\begin{table}[ht!]
\caption{The number of samples in the training, validation, and test sets for the CelebA dataset.}
\label{tab:celebA_samples}
\centering
\small
\begin{tabular}{c|ccc}
\toprule
Split & Train & Validation & Test \\
\midrule
(NonBlond, female) & 12,426 & 8,535 & 9,767 \\
\midrule
(NonBlond, male) & 11,841 & 8,276 & 7,535 \\ 
\midrule
(Blond, female) & 22,880 & 2,874 & 2,480 \\
\midrule
(Blond, male) & 1,387 & 182 & 180 \\
\bottomrule
\end{tabular}
\end{table}
\subsubsection{CelebA*}
CelebA is a face-attribute dataset where the task is to classify hair color as \emph{blond} or \emph{non-blond}, with gender acting as the spurious attribute. In the original CelebA setting, the main underrepresented group is \emph{(blond, male)}, while the other groups are comparatively much larger. For our experiments, we construct a controlled subsampled version, denoted as CelebA*, for two reasons. First, we aim to create a setting with \emph{two} minority groups rather than only one, which makes the benchmark more consistent with our group-robustness setting. Second, CelebA is large, and computing CLIP embeddings for the full dataset would incur substantial computational cost.

Specifically, we subsample the training set to contain 5,000 examples from \emph{(blond, female)} and \emph{(non-blond, male)}, and 100 examples from \emph{(blond, male)} and \emph{(non-blond, female)}. For validation and test, we sample 100 examples from each group. As shown in Table~\ref{tab:celeba_star_samples}, this construction yields a training split with two majority groups and two minority groups, while keeping the evaluation splits balanced across groups. 
\begin{table}[ht!]
\caption{The number of samples in the training, validation, and test sets for the CelebA* dataset.}
\label{tab:celeba_star_samples}
\centering
\small
\begin{tabular}{c|ccc}
\toprule
Split & Train & Validation & Test \\
\midrule
(Blond, female) & 5,000 & 100 & 100 \\
\midrule
(Non-blond, male) & 5,000 & 100 & 100 \\
\midrule
(Blond, male) & 100 & 100 & 100 \\
\midrule
(Non-blond, female) & 100 & 100 & 100 \\
\bottomrule
\end{tabular}
\end{table}
\subsection{Dominoes}
Dominoes is a synthetic dataset introduced in \cite{dominoes}. Each image is composed of two parts: a CIFAR-10 image, either \emph{truck} or \emph{car}, which determines the class label, and an MNIST digit, either \emph{0} or \emph{1}, which serves as the spurious attribute. The MNIST digit is placed below the CIFAR image and is correlated with the target label. As shown in Table~\ref{tab:dominoes_samples}, the training split is imbalanced across groups, while the validation and test splits are nearly balanced. This setup makes Dominoes a useful benchmark for evaluating robustness under spurious correlations.

\subsubsection{DDB \cite{ddb} on Dominoes}
\begin{figure}
    \centering
    \includegraphics[width=0.5\linewidth]{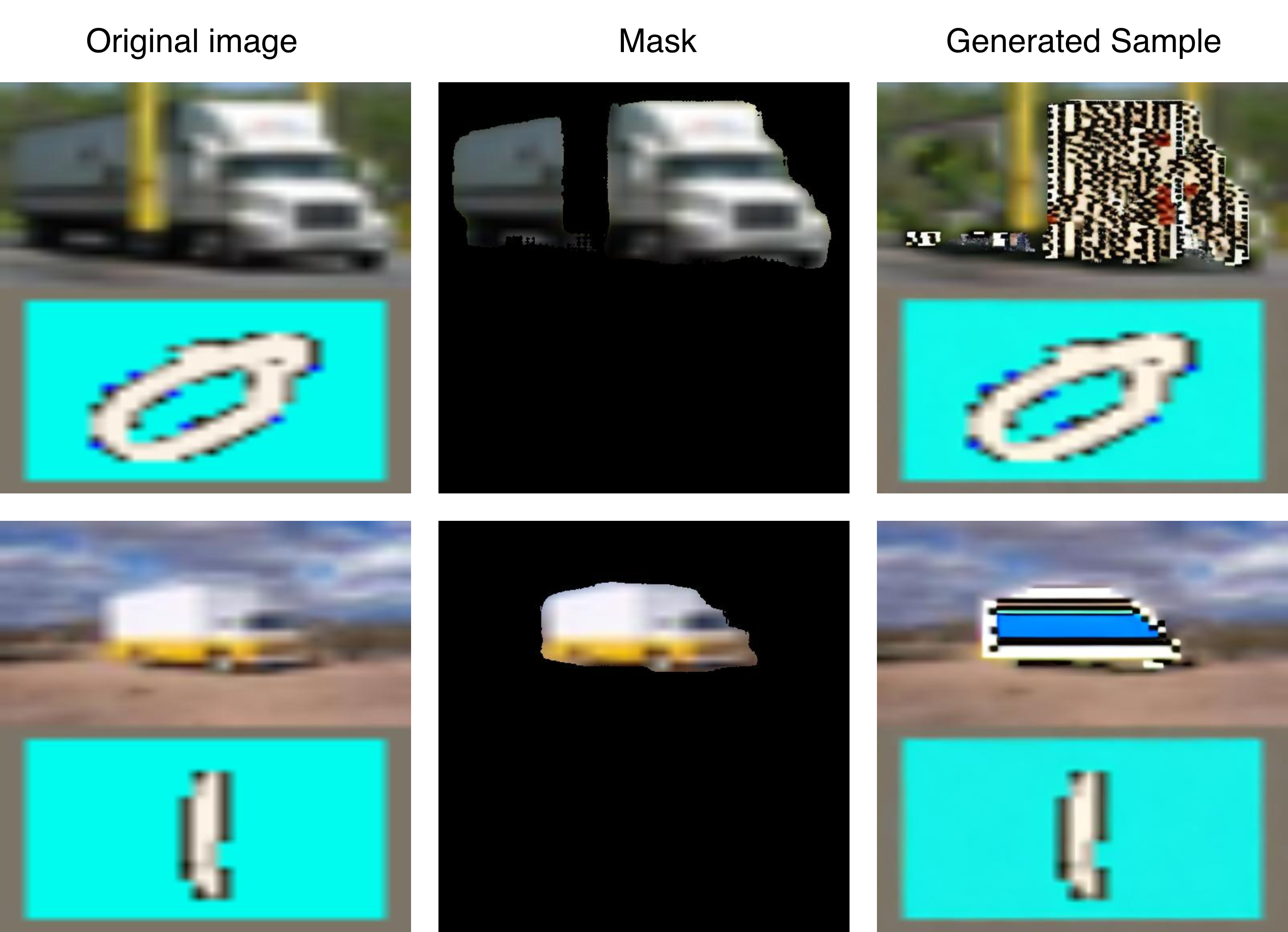}
    \caption{DDB generated samples on Dominoes dataset}
    \label{fig:ddb}
\end{figure}

As noted in \cite{ddb}, the performance of DDB is inherently constrained by the capabilities of the underlying diffusion model. This limitation is particularly evident on the Dominoes dataset, where the images are synthetic and low-resolution, making it difficult for diffusion-based generation to reliably introduce minority-group features. In contrast, HSFM does not rely on auxiliary generative models for data augmentation. Instead, it operates directly in the latent feature space, which avoids this difficulty. Qualitative examples are provided in Figure~\ref{fig:ddb}

\begin{table}[ht!]
\caption{The number of samples in the training, validation, and test sets for the Dominoes dataset.}
\label{tab:dominoes_samples}
\centering
\small
\begin{tabular}{c|ccc}
\toprule
Split & Train & Validation & Test \\
\midrule
(car, 0)   & 3,592 & 500 & 490 \\
\midrule
(car, 1)   & 398 & 499 & 490 \\
\midrule
(truck, 0) & 398 & 499 & 490 \\
\midrule
(truck, 1) & 3,592 & 500 & 490 \\
\bottomrule
\end{tabular}
\end{table}

\subsection{Metashift}
\vspace{1mm}
\noindent\textit{MetaShift. \cite{metashift}}
MetaShift is a distribution-shift benchmark for classifying images of \emph{cats} and \emph{dogs}. Following \cite{DISC}, we construct the training set using images of \emph{cats on sofas}, \emph{cats on beds}, \emph{dogs on benches}, and \emph{dogs on bikes}. We then evaluate the model on images of cats and dogs with \emph{shelf} backgrounds, which are not observed during training. As shown in Table~\ref{tab:metashift_samples}, the training and evaluation splits contain different background contexts, making this benchmark suitable for assessing robustness under distribution shift.

\begin{table}[ht!]
\caption{The number of samples in the training, validation, and test sets for the MetaShift dataset.}
\label{tab:metashift_samples}
\centering
\small
\begin{tabular}{c|ccc}
\toprule
Split & Train & Validation & Test \\
\midrule
(Cat, sofa) & 231 & 0 & 0 \\
\midrule
(Cat, bed) & 380 & 0 & 0 \\ 
\midrule
(Dog, bench) & 145 & 0 & 0 \\
\midrule
(Dog, bike) & 367 & 0 & 0 \\
\midrule
(Cat, shelf) & 0 & 34 & 201 \\
\midrule
(Dog, shelf) & 0 & 47 & 259 \\
\bottomrule
\end{tabular}
\end{table}\vspace{1mm}
\noindent\textit{MetaShift. \cite{metashift}}
MetaShift is a distribution-shift benchmark for classifying images of \emph{cats} and \emph{dogs}. Following \cite{DISC}, we construct the training set using images of \emph{cats on sofas}, \emph{cats on beds}, \emph{dogs on benches}, and \emph{dogs on bikes}. We then evaluate the model on images of cats and dogs with \emph{shelf} backgrounds, which are not observed during training. As shown in Table~\ref{tab:metashift_samples}, the training and evaluation splits contain different background contexts, making this benchmark suitable for assessing robustness under distribution shift.

\begin{table}[ht!]
\caption{The number of samples in the training, validation, and test sets for the MetaShift dataset.}
\label{tab:metashift_samples}
\centering
\small
\begin{tabular}{c|ccc}
\toprule
Split & Train & Validation & Test \\
\midrule
(Cat, sofa) & 231 & 0 & 0 \\
\midrule
(Cat, bed) & 380 & 0 & 0 \\ 
\midrule
(Dog, bench) & 145 & 0 & 0 \\
\midrule
(Dog, bike) & 367 & 0 & 0 \\
\midrule
(Cat, shelf) & 0 & 34 & 201 \\
\midrule
(Dog, shelf) & 0 & 47 & 259 \\
\bottomrule
\end{tabular}
\end{table}

\subsection{CUB-200-2011}
The Caltech-USCD Birds-200-2011 (CUB-200-2011) dataset \cite{Birds} is a famous benchmark for fine-grained image classification. It contains 11,788 images of 200 birds species collected from North America. The dataset is split into 5,994 training images and 5,794 test images. Each image is annotated with detailed metadata including bounding boxes, part locations, and attribute annotations. Due to the high inter-class similarity and large intra-class variations among birds, CUB-200-2011 presents significant challenges for fine-grained recognition models.

\subsection{Oxford Flowers}
The Oxford-102 Flowers dataset \cite{Flowers} contains 8,189 images from 102 flower categories  commonly found in the United Kingdom. Each class has between 40 and 258 images, capturing large variation in scale, pose, and lighting conditions. The dataset provides predefined splits consisting of 1,020 training images, 1,020 validation images, and 6,149 test images. It is commonly used to evaluate fine-grained classification methods due to subtle visual differences among flower species.

\subsection{Stanford Cars}
The Stanford Cars dataset \cite{Cars} is a fine-grained vehicle classification benchmark containing 16,185 images of 196 classes of cars. The classes correspond to different cats makes models and production years. It is split into 8,144 training images and 8,041 test images. Images are collected from real-world scenes and include major variations in occlusion, viewpoint, background clutter, making the dataset challenging for fine-grained visual recognition.

\section{Experiments}
\subsection{Spurious Correlation}

\subsubsection{Hyperparameters}

Detailed hyperparameters for all experiments are provided in Tables~\ref{tab:resnet_hyperparameters}, \ref{tab:vit_hyperparameters}, and \ref{tab:convnext_hyperparameters}. For all settings, the learning rate used for the ERM updates is the same as the inner-loop learning rate.

\begin{table*}[t]
\centering
\caption{Hyperparameters used for HSFM with a ResNet backbone on each dataset.}
\label{tab:resnet_hyperparameters}
\small
\begin{tabular}{lcccc}
\toprule
Hyperparameter & CelebA & Waterbirds & MetaShift & Dominoes \\
\midrule
Support set size per class & 1024 & 16 & 32 & 16 \\
$T$ & 10 & 15 & 15 & 10 \\
Inner learning rate & $5 \times 10^{-5}$ & $5 \times 10^{-5}$ & $5 \times 10^{-5}$ & $1 \times 10^{-4}$ \\
Outer learning rate & $1 \times 10^{-1}$ & $1$ & $1 \times 10^{-2}$ & $1$ \\
$K_{H}$ & 5 & 15 & 10 & 15 \\

$K_{hard}$ & 256 & 64 & 64 & 256 \\
epochs & 20 & 40 & 10 & 40 \\
\bottomrule
\end{tabular}
\end{table*}

\begin{table*}[t]
\centering
\caption{Hyperparameters used for HSFM with a ViT backbone on each dataset.}
\label{tab:vit_hyperparameters}
\small
\begin{tabular}{lcccc}
\toprule
Hyperparameter & CelebA & Waterbirds & MetaShift & Dominoes \\
\midrule
Support set size per class & 1024 & 128 & 64 & 128 \\
 $T$ & 10 & 10 & 10 & 10 \\
Inner learning rate & $1 \times 10^{-4}$ & $1 \times 10^{-4}$ & $1 \times 10^{-5}$ & $1 \times 10^{-4}$ \\
Outer learning rate & $1$ & $1$ & $1 \times 10^{-2}$ & $1$ \\
$K_{H}$ & 15 & 15 & 10 & 15 \\

$K_{Hard}$ & 256 & 32 & 8 & 256 \\
epochs & 50 & 40 & 30 & 40 \\
\bottomrule
\end{tabular}
\end{table*}

\begin{table*}[t]
\centering
\caption{Hyperparameters used for HSFM with a ConvNeXt backbone on each dataset.}
\label{tab:convnext_hyperparameters}
\small
\begin{tabular}{lcc}
\toprule
Hyperparameter & CelebA & Waterbirds \\
\midrule
Support set size per class & 32 & 128 \\
$T$ & 10 & 10 \\
Inner learning rate & $1 \times 10^{-4}$ & $1 \times 10^{-4}$ \\
Outer learning rate & $1$ & $1$ \\
$K_{H}$ & 15 & 15 \\

$K_{Hard}$ & 8 & 16 \\
epochs & 40 & 40 \\
\bottomrule
\end{tabular}
\end{table*}

\subsubsection{Computational Cost}

Table~\ref{tab:computational_cost} reports the total runtime of HSFM after ERM training on each dataset, measured on a single NVIDIA A100 GPU. The results show that our method is computationally efficient, with all experiments completing within a few minutes. For direct comparison on the same hardware, Table~\ref{tab:waterbirds_runtime_comparison} compares our runtime on Waterbirds against prior methods. The runtimes for DFR, CnC, JTT, MaskTune, and DaC are taken from DaC \cite{dac}, while the runtime for our method is measured in our implementation. Our method is faster than all compared baselines. We also note that diffusion-based augmentation methods can incur substantially higher cost; for example, DDB \cite{ddb} reports that image generation alone requires around one hour.

\begin{table}[t]
\centering
\caption{Training time comparison on the Waterbirds dataset, excluding ERM training, measured on an NVIDIA A100 GPU. Runtimes for DFR, CnC, JTT, MaskTune, and DaC are reported from DaC \cite{dac}, and our runtime is measured in our implementation.}
\label{tab:waterbirds_runtime_comparison}
\small
\begin{tabular}{lcccccc}
\toprule
Method & DFR & CnC & JTT & MaskTune & DaC & Ours \\
\midrule
Time (min) & 4 & 85 & 58 & 6.5 & 18.9 & 1.43 \\
\bottomrule
\end{tabular}
\end{table}

\begin{table}[t]
\centering
\caption{Total program runtime of HSFM on each dataset after ERM training, measured on a single NVIDIA A100 GPU.}
\label{tab:computational_cost}
\small
\begin{tabular}{lc}
\toprule
Dataset  & Runtime (minutes) \\
\midrule
CelebA     &  2.98 \\
Waterbirds &  1.43 \\
Dominoes   &  0.26 \\
MetaShift  & 0.27 \\
\bottomrule
\end{tabular}
\end{table}

\subsection{Fine Grained Classification}

\subsubsection{Hyperparameters}
For all experiments, we adopt an ResNet-50 backbone as the feature extractor. We resized input images 224x224 and normalized using standard ImageNet statistics. The classifier head is optimized using Adam Optimizer with a learning rate $1\times10^{-3}$ for head fine-tuning and $5\times10^{-5}$ for meta-optimization steps. We use a batch size of 128 for standard fine-grained classification experiments and 256 for the meta-optimization setting. Training is performed for 150 epochs on the CUB-200, Oxford-102 Flowers, and Stanford Cars datasets. To address class imbalance, we optionally apply inverse-frequency class reweighting in the loss function. The validation set is split using a $50/50$ partition for training and evaluation during model selection.
\par
For the meta-optimization stage, we initialize 16 latent feature vectors per class and run 40 meta-iterations. Each iteration consists of 15 inner optimization steps followed by 15 ERM updates of the classifier. Hard validation samples are selected using a top-$k$ loss-based sampling strategy with k=64
per class to focus on challenging during optimization. Gradient clipping maximum norm of 10 is applied to stabilize training. All experiments are implemented in PyTorch and executed on a workstation equipped with an NVIDIA RTX Pro Blackwell GPU 96GB VRAM. Backbone features are cached to avoid redundant computation during meta-optimization significantly reducing training overhead.


\subsubsection{Computational Cost}

We analyze the computational efficiency of our method across three fine-grained classification datasets: Stanford-Cars \cite{Cars}, CUB-Birds \cite{WahCUB_200_2011}, and Oxford-Flowers102 \cite{Flowers}. For a fair comparison, all experiments were conducted on an NVIDIA RTX Pro Blackwell GPU with 96GB VRAM. Under this setup, training a standard ERM model requires approximately 2 hours, whereas our method completes within 10 minutes. This represents a substantial reduction in computational cost while maintaining state-of-the-art performance.

\bibliographystyle{splncs04}
\bibliography{main}

\end{document}